\documentclass{article}

\usepackage{resources/PRIMEarxiv}

\usepackage[utf8]{inputenc} 
\usepackage[T1]{fontenc}    
\usepackage{hyperref}       
\usepackage{url}            
\usepackage{booktabs}       
\usepackage{amsfonts}       
\usepackage{nicefrac}       
\usepackage{microtype}      
\usepackage{lipsum}
\usepackage{fancyhdr}       
\usepackage{graphicx}       
\graphicspath{{media/}}     
\usepackage{amsmath,amsthm,amsfonts,amssymb,amsopn,cancel,xcolor}
\usepackage{multirow, booktabs}
\usepackage{subcaption}

\title{ViTO: Vision Transformer-Operator 
}

\author{
  Oded Ovadia, Eli Turkel \\
  Department of Applied Mathematics \\
  Tel Aviv University \\
  Tel Aviv \\
  \texttt{odedovadia@mail.tau.ac.il, eliturkel@gmail.com } \\
  \And
  Adar Kahana \\
  Division of Applied Mathematics \\
  Brown University \\
  Providence, RI \\
  \texttt{adar\_kahana@brown.edu}  \\
    \And
  George Em Karniadakis \\
  Division of Applied Mathematics \\
  Brown University \\
  Providence, RI \\
  and\\Advanced Computing, Mathematics and Data Division \\
  Pacific Northwest National Laboratory \\
  Richland, WA \\
  \texttt{george\_karniadakis@brown.edu}  \\
  \And
  Panos Stinis \\
  Advanced Computing, Mathematics and Data Division \\
  Pacific Northwest National Laboratory \\
  Richland, WA \\
  \texttt{panagiotis.stinis@pnnl.gov} \\
}

\begin{document}
\maketitle

\begin{abstract}
We combine vision transformers with operator learning to solve diverse inverse problems described by partial differential equations (PDEs). Our approach, named ViTO, combines a U-Net based architecture with a vision transformer. We apply ViTO to solve inverse PDE problems of increasing complexity, namely for the wave equation, the Navier-Stokes equations and the Darcy equation. We focus on the more challenging case of super-resolution, where the input dataset for the inverse problem is at a significantly coarser resolution than the  output. The results we obtain are comparable or exceed the leading operator network benchmarks in terms of accuracy. Furthermore, ViTO's architecture has a small number of trainable parameters (less than 10\% of the leading competitor), resulting in a performance speed-up of over 5x when averaged over the various test cases.


\end{abstract}


\section{Introduction}

Operator learning refers to training neural networks to represent mappings between families of functions. For example, if we want to infer the acoustic wave pressure in the ocean for each and every initial source, we can define the operator as the mapping from the initial source (a function, initial pressure in every point of the domain) to the pressures at a later time (also a function, future pressures in every point of the domain). The main advantage of operator learning is that, after the operator has been learned, no further training is needed and the solution, e.g., for a partial differential equation (PDE), which can be very expensive using classic methods, can be simply inferred (estimated) by the network with negligible computational cost in real-time. The first operator learning method was introduced by Lu \textit{et al.}, named the Deep Operator Network (DeepONet) \cite{deeponet}. DeepONet is composed of a branch and a trunk; the branch learns the input function space of the operator, while the trunk learns the space of functions onto which the output is projected. By multiplying the branch with the trunk, the projection provides a representation of the output function space. Another popular invention is the Fourier Neural Operator \cite{FNO, FNO2}, which is based on replacing the kernel integral operator with a convolution operator defined in Fourier space by employing a fast Fourier transform on the input space. It uses a ResNet but does not have a trunk net.

When modeling a physical system/experiment we usually find one of two scenarios. The first scenario is the \textit{forward problem}, when given a set of conditions, one attempts to simulate the physical process. For example, given an initial source, find the acoustic wave amplitude in the ocean after some time. The second scenario is the opposite one, called the \textit{inverse problem}. Given the state of the physical experiment, find the causal condition that led to that state. For example, given measurements of the acoustic pressures in the ocean at some time instant, find the source that started emitting the acoustic sounds. The inverse problem is often considered more challenging since one has access to limited data (for example, recordings at a small set of sensors only), the data may be noisy and of low resolution, some recordings may be missing, etc. Inverse problems are often ill-posed, meaning they do not necessarily have a solution and if so, it may not necessarily be unique. In this work, we focus on inverse problems given {\em relatively sparse} data sets. 


A recent innovation in the field of deep learning is the so-called transformers \cite{transformer}, which refer to deep neural networks that have an attention mechanism. The attention module attempts to understand context from the given input. To learn the context, the attention mechanism operates on a discrete embedding of the data that is composed of tokens. An immediate example is Natural Language Processing (NLP) using transformers, where a sentence is embedded using tokens according to a specific vocabulary. A challenge that arises when using transformers is considering non-discrete data. For example, instead of sentences (sequence of words), we would like to embed a continuous function or signal. The literature offers methods to achieve that, as introduced in the following section. One prominent method, explored in the current work, is called Vision Transformers (ViT) \cite{vit}. The vision transformers receive an image as input, for example the initial condition of a system (in the forward problem), or a future state of a system (in the inverse problem). Then, the image is split into small regions, often referred to as patches, and each region acts as a token. The vision transformer extracts the context from the tokens (regions of the image), and thus is able to utilize the attention mechanism for the continuous signal and make accurate predictions. In the original ViT paper \cite{vit}, the authors demonstrated how ViT outperforms the state-of-the-art (SOTA) methods for image classification, including ImageNet \cite{imagenet}, CIFAR \cite{cifar}, Oxford pets and flowers \cite{oxford_pets}, and VTAB \cite{VTAB}. In addition, the benchmark ViT model is up to four times more efficient than the SOTA methods used as reference. In \cite{okolo2022ievit}, Okolo \textit{et. al.} used a ViT for X-ray image classification, while in \cite{chen2021vit} it was used for unsupervised volumetric medical image registration. ViTs are  currently being adopted in various areas of research.

In the current work, we are interested in using transformers for operator learning. There have been some recent attempts to use various types of transformers for operator learning \cite{transformerPDE1, transformerPDE2, transformerPDE3, transformer_cao_choose, transformer_cao2}. For example, in \cite{transformerPDE1}, Li \textit{et. al.} use transformers to approximate \emph{forward} solutions of PDEs with operator learning. They propose an innovative way for choosing the collocation points and iterate through time to find the solution at those points. The attention mechanism is split in such a way that the query, key and value are split into different parts of the forward pass and is implemented as a multi-layer perceptron (MLP).  The latent encoding, which is the outcome of the combination of the three components, is the embedding of the coordinates used for the spatio-temporal input of the PDE. In \cite{transformer_cao_choose}, Cao presents a method to combine FNO with attention to improve the performance for PDE solutions, by replacing the softmax (often used in transformers especially in classification) with a linear variant of it that does not involve normalization.

Here, we introduce a novel way to perform operator learning using vision transformers combined with a U-net \cite{ronneberger2015u, chen2021transunet} based architecture to design the Vision Transformer-Operator or ViTO. We apply ViTO to solve inverse problems of increasing complexity, obtaining the solution at high resolution using only sparse and low resolution data. Compared to SOTA results for operator learning, our current results exceed the leading operator network benchmarks in terms of accuracy, and they are also obtained at a significant speedup.

The paper is organized as follows. Section \ref{method} presents the proposed methodology. Section \ref{results} presents numerical results for a collection of inverse problems of increasing complexity. Section \ref{discussion} offers a discussion of the results and directions for future work.

\section{Methodology}\label{method}

\subsection{Mathematical formulation of operator learning}

We first present the general problem formulation of operator learning for PDEs before focusing on the particular case of inverse problems. We follow the DeepONet theory and notation, as given by \cite{deeponet, deeponet-fno}.

\subsubsection{PDE operators}\label{PDE_operators}
Typically, when tackling a forward PDE problem, our objective is to determine the solution to the PDE. Thus, we aim to approximate the PDE solution by utilizing a set of parameters that describe the PDE problem setup. Such parameters include initial and boundary conditions, forcing terms, and other physical characteristics that may vary between different PDEs. Hence, forward PDE problems can be formulated as a mapping between an input function that corresponds to these parameters to an output function representing the solution. 

Mathematically, let $v$ denote the input function defined on some physical domain $D \in \mathbb{R}^d$ and $u$ denote the corresponding output function defined on the physical domain $D' \in \mathbb{R}^{d'}$:
 \begin{align*}
     v: D \ni x \longmapsto v(x) \in \mathbb{R} , \\
     u: D' \ni \xi \longmapsto u(\xi) \in \mathbb{R} .
 \end{align*}

Let $\mathcal{V}$ and $\mathcal{U}$ be the spaces of the functions $v$ and $u$, respectively. Then, the mapping from $v$ to $u$ is defined by an operator $\mathcal{G}$:

\begin{equation*}
     \mathcal{G}: \mathcal{V} \ni v \longmapsto u \in \mathcal{U} .
\end{equation*}

This operator describes the forward problem. For example, in many applications $v$ is the initial condition of the PDE and $u$ is its solution at some final time. However, in this work we are interested in the inverse problem. So, the operator corresponding to the inverse problem is of the form:

 \begin{equation}\label{inverse_operator}
     \Tilde{\mathcal{G}}: \mathcal{U} \ni u \longmapsto v \in \mathcal{V} .
\end{equation}

Continuing the previous example, the relevant inverse problem would be to retrieve the initial condition of the PDE given a snapshot of its solution.

We note that in many cases, the inverse operator relates to an ill-posed problem \cite{hadamard1902problemes}. This type of problem is generally considered more challenging, particularly when dealing with incomplete or noisy data.

\subsubsection{Super-resolution}\label{SR}

For most applications, it is impossible to get the full analytical solution of a PDE. In some cases, it is even hard to get a discrete approximation of the solution on a fine mesh due to computational, physical, or experimental difficulties. This is especially common in the domain of inverse problems, where the input function is often derived from sensor measurements of physical phenomena. These considerations often lead to a low-resolution mesh for the discrete approximation.

Low-resolution data is challenging to use. The main goal of Super-Resolution (SR) methods is to produce high-resolution accurate results given low-resolution input data. In this work, we do not treat the SR aspect as a separate problem. Instead, we combine it with the inverse operator defined in the previous section \ref{PDE_operators} to form a unified inverse-SR framework. Using the same notation as before, instead of getting functions $u \in \mathcal{U}, v \in \mathcal{V}$, we get discrete approximations of these functions on meshes. However, $u \in \mathcal{U}$ is discretized using a much coarser mesh in comparison to $v \in \mathcal{V}$. For example, the input might be a low-resolution snapshot of a PDE solution, while the desired output would be a high resolution discretization of the initial condition.

\subsection{Data driven formulation of operator learning}\label{data_driven}

The goal is to approximate the operator $\Tilde{\mathcal{G}}$ in \eqref{inverse_operator}, when the input is in low-resolution and the output is in high-resolution, using a ViT-based neural network. We define a dataset, where each sample is composed of pairs of discretized functions: 
$\mathcal{T} = \{(\mathrm{u}^{(1)}, \mathrm{v}^{(1)}), (\mathrm{u}^{(2)}, \mathrm{v}^{(2)}), \hdots, (\mathrm{u}^{(N)}, \mathrm{v}^{(N)})\}$ such that $\forall n,  \mathrm{u}^{(n)}$ and $\mathrm{v}^{(n)}$ are the projections of functions  ${u} ^{(n)} \in \mathcal{U}$ and ${v} ^{(n)}\in \mathcal{V}$ onto discrete meshes $\mathcal{M}_{u}$ and $\mathcal{M}_{v}$, respectively. We let $\mathcal{M}_{u}$ and $\mathcal{M}_{v}$ remain constant for all samples, and assume that the discretization is equispaced. For ease of notation and without loss of generality, we assume that the domain coordinates are positive. Then, in the two-dimensional case, which is used for all of the numerical experiments, these meshes are written as:

\begin{align*}
    \mathcal{M}_{u} = \{ (i \Delta_{x, u}, \: \: j \Delta_{y, u}) \: \: | \: \: i,j \in \mathbb{N}, i \leq N_{x, u}, j \leq N_{y, u}  \} \\ 
    \mathcal{M}_{v} = \{ (i \Delta_{x, v}, \: \: j \Delta_{y, v}) \: \: | \: \: i,j \in \mathbb{N}, i \leq N_{x, v}, j \leq N_{y, v}  \},
\end{align*}

where $\Delta_{x, \cdot}, \Delta_{y, \cdot}, N_{x, \cdot}, N_{y, \cdot}$ determine the resolution of the discretization. We use $\# \mathcal{M}_{u} = N_{x, u} \cdot N_{y, u}$ to note the number of points in the set $\mathcal{M}_{u}$.

If $\mathcal{M}_{u}$ and $\mathcal{M}_{v}$ have the same number of points ($\# \mathcal{M}_{u} = \# \mathcal{M}_{v}$.), SR is not performed, and instead, we are describing a standard inverse problem. However, as highlighted in the previous section \ref{SR}, the focus is on scenarios where the input grid is significantly coarser than the output grid, i.e., $\# \mathcal{M}_{u} \ll \# \mathcal{M}_{v}$. To achieve this, we can choose a super-resolution factor $s > 1$ that specifies the relationship between the discretization. We choose $N_{\cdot, v}$ such that $N_{x, v} = s N_{x, u}$ and $ N_{y, v} = s N_{y, u}$ to attain a SR factor of $s$. This results in the output grid size becoming $\# \mathcal{M}_{v} = s^2 N_{x, u} N_{y, u}$. For large values of $s$, this renders the problem considerably more challenging.

\subsection{Network architecture}\label{network_arch}

\begin{figure}[htb]
\includegraphics[scale=0.45]{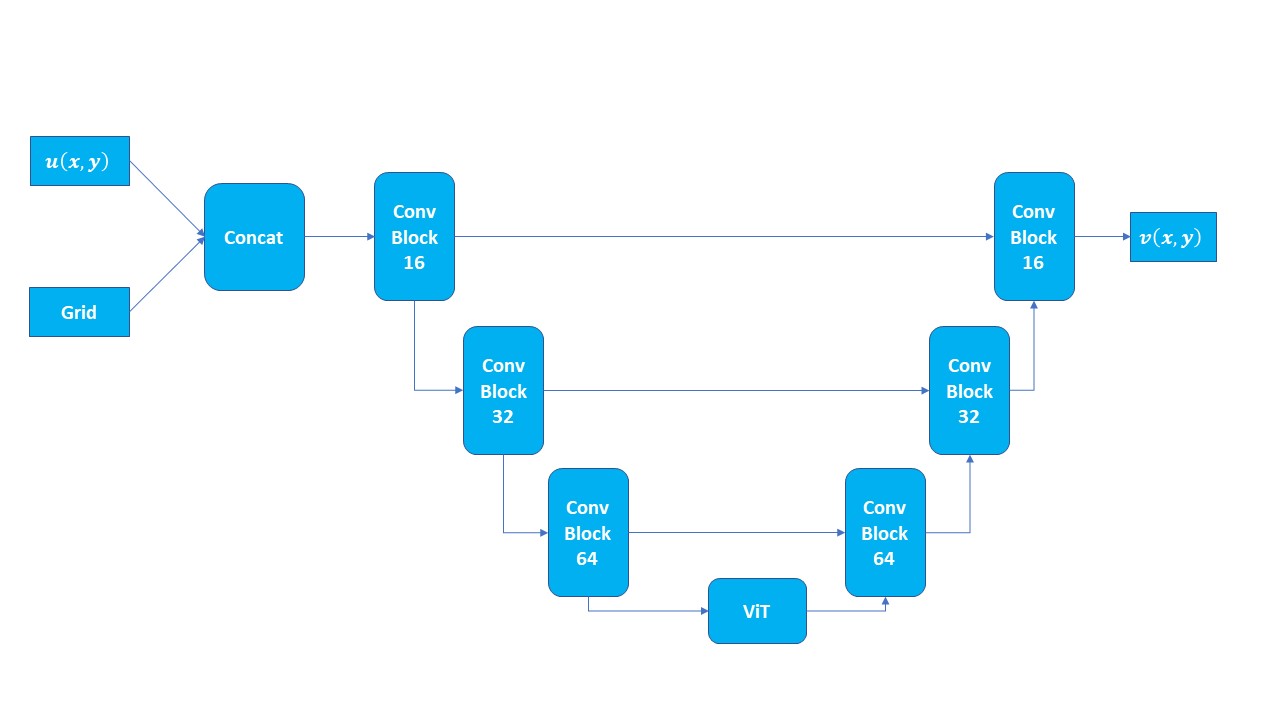}
\centering
\caption{The architecture of the ViTO deep neural network. The inputs are the discretized function $u(x,y)$ and the grid points, concatenated and inserted into U-net convolutional blocks. In the lowest level of the U-net the ViT is employed.}
\label{fig:architecture}
\end{figure}

A modified version of the TransUNet \cite{chen2021transunet} architecture is employed in this study, wherein a U-Net \cite{ronneberger2015u} backbone is integrated with a ViT (see Figure \ref{fig:architecture}). The U-Net model comprises an encoder-decoder structure with interconnecting skip connections. The U-Net architecture has emerged as a powerful technique in the computer vision field, particularly in the realm of segmentation problems. Given the image-to-image nature of the data-driven problem outlined in \ref{data_driven}, utilizing segmentation tools is a natural choice.

\begin{figure}[htb]
\includegraphics[scale=0.45]{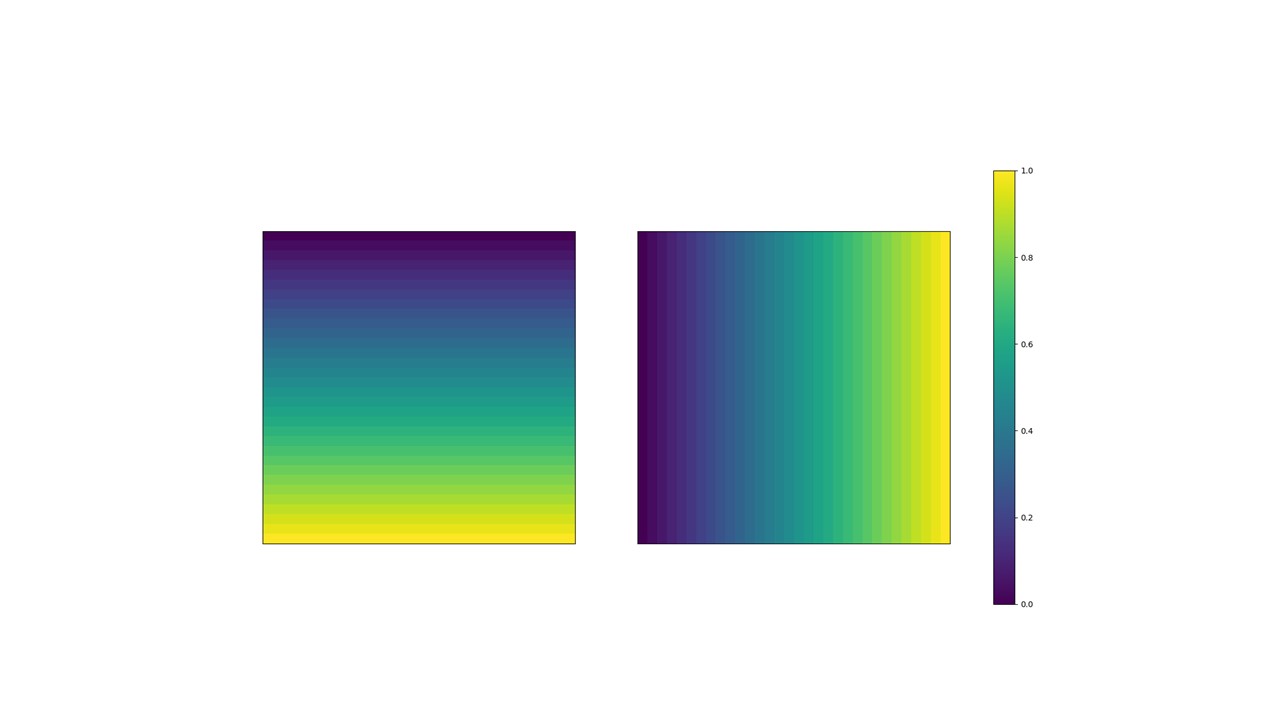}
\centering
\caption{An example of discrete $X, Y$ grid encoded inputs for a problem defined in $[0, 1]^2$.}\label{fig:grids}
\end{figure}

The network has two inputs: the observed solution of the PDE $u$, and its corresponding numerical grid $\mathcal{M}_{u}$. In the two-dimensional case, $u$ is represented by a two-dimensional matrix, where all elements are values of the $u$ at points on the grid  $\mathcal{M}_{u}$. We want the network to be exposed to the grid itself, so it can learn some relation between the $(x,y)$ values of a grid point and their corresponding solution value $u(x,y)$. We achieve this by a simple encoding of the grid using two matrices representing discretizations in the $x$ and $y$ directions, as seen in Figure \ref{fig:grids}. The $i$-th row of the $x$-matrix is defined as a vector consisting of $x_i = i \Delta_{x, u}$ repeated $N_{y, u}$ times. Similarly, the $j$-th column of the $y$-matrix is defined as a column vector consisting of $y_j = j \Delta_{y, u}$ repeated $N_{x, u}$ times. We also use bilinear interpolation to modify the sizes of the inputs to the desired output shape. We make sure that the results of the interpolation operation is divisible by $16$, so it could be compatible with the downsampling of the U-Net.

The U-Net architecture we employ is composed of three convolutional blocks for both the encoder and decoder (see Figure \ref{fig:architecture}). Each block is comprised of three convolutional layers, equipped with a residual skip connection \cite{ResNet}. All convolutions are followed by a batch normalization \cite{batch_norm} layer and a GELU activation function \cite{hendrycks2016gaussian}. The final layer of each block performs either a downsampling operation for the encoder or an upsampling operation for the decoder.

In our approach, we utilize a ViT within the latent space of the U-Net by taking the encoded values as input. As the encoding is significantly smaller than the original inputs due to the U-Net's downsampling nature, we can use a patch size of $1 \times 1$ without any computational difficulties.

The original ViT utilizes absolute positional embedding via a linear projection layer. However, this can be problematic in PDE applications as it requires all inputs to have the same shape. In operator learning, we are often interested in models that can handle inputs of various sizes. To address this issue, we employ a form of relative conditional embedding \cite{shaw2018self, wu2021rethinking}, where we use convolutions to learn the relationships between tokens instead of linearly projecting their absolute positions within the representation in the latent space. Specifically, we use a separable convolutional layer \cite{chollet2017xception} followed by a standard convolutional layer, similar to the approach proposed in \cite{chu2021conditional}. 

\subsection{Training loss function}

The loss function is defined as the mean relative $\textit{L}^2$ error:

\begin{align}
    \label{eqn:loss}
    \mathcal{L}=\frac{1}{N} \sum_{j=1}^N\frac{||\mathrm{\hat{v}}^{(j)}-\mathrm{v}^{(j)}||_2}{\varepsilon+||\mathrm{v}^{(j)}||_2},
\end{align}

where $N$ is the size of training data, $\mathrm{v}^{(j)}$ is the $j$-th ground-truth sample of the training data, $\mathrm{\hat{v}}^{(j)}$ is the $j$-th sample prediction, and $\varepsilon$ is a small number to  prevent a zero denominator and stabilize the loss. Note that the inputs and outputs of the model are two-dimensional, so they are flattened inside the loss function. 

\section{Numerical results}\label{results}

We apply the ViTO method on various ill-posed two-dimensional inverse problems. The conducted tests include three PDEs: the acoustic wave equation, time-dependent incompressible Navier-Stokes, and a steady-state Darcy flow equation. In all cases we compare the results of the ViTO to three other popular methods in the scientific machine learning literature: 1) DeepONet \cite{deeponet}, 2) FNO \cite{FNO}, and 3) a standard ResNet \cite{ResNet}. For each method we compute the relative $\textit{L}^2$ error compared to the ground truth data.  We also measure relevant information regarding the training time and the number of parameters of used by each method. Details regarding the data generation process can be found in each of the following sections.

An important consideration in the domain of inverse problems is robustness to noise. Since inverse problems are often ill-posed, even a small amount of noise in the observed data can greatly amplify the numerical error \cite{ill_posed}. To test how well ViTO can handle noise, we run all experiments twice: with noise and without noise. In both cases we train the model with the relevant amount of noise. We used zero-mean Gaussian additive noise, which is a  common choice. Since different PDEs can behave quite differently, we make sure that the variance of the Gaussian noise is dependent on the input data. The operation of adding noise is given by:
$\mathrm{D} \ni x^{(n)} \longmapsto x^{(n)} + \gamma \mathcal{N}(0, \sigma^2_{\mathrm{D}})$
where $x^{(n)}$ is an input sample in the dataset $\mathrm{D}$, $\sigma^2_{\mathrm{D}}$ is the variance of the entire dataset, and $\gamma$ is the desired noise level, e.g. $\gamma = 0.1$ is equivalent to $10 \%$ noise.

For all our evaluations, we utilized a super-resolution scale factor of 8, which means, for example, that an image of size $16 \times 16$ would be mapped to an image of size $128 \times 128$. It should be noted that a magnification of $\times 8$ is considerably high. Although lower SR factors are utilized in several benchmarking datasets, the choice of $\times 8$ is still prevalent for certain datasets such as Urban100 \cite{urban100} and the Berkeley Segmentation Dataset \cite{BSD}.

For the FNO we used the same architecture as described in \cite{FNO} for the FNO-2D network: four Fourier layers with width of 32 and 12 modes. Each layer was followed by a GeLU \cite{hendrycks2016gaussian} activation function. For the ResNet we used 3 residual blocks with 16, 32, and 64 filters for each convolution within each block of depth 3. In the DeepONet case we used the ResNet described before as the branch network, 4 hidden fully connected layers for the trunk network, and 256 neurons for the latent dimension.

Following standard machine learning practice, we split all datasets into train, test, and validation sets. During training, we monitor the relative $\textit{L}^2$ losses and save the model with the lowest validation loss. Unless stated otherwise, all models are trained with a batch size of 100 for 500 epochs, subject to an early stopping criterion of 50 consecutive epochs with no validation loss improvement. The optimizer of choice is the Adam/AdamW optimizer \cite{kingma2014adam, adamw} with an initial learning rate $10^{-3}$ and weight decay $10^{-4}$. The learning rate is updated throughout the training process using cosine annealing \cite{loshchilov2016sgdr}. The main code was implemented in PyTorch \cite{pytorch}, and the DeepONet was implemented using DeepXDE \cite{deepxde} with a PyTorch backend. All computations were conducted using a single RTX-4090 GPU.

\begin{table}[!ht]
\centering
\begin{tabular}{|c|c|c|c|}
\hline
Method       & $\#$ of parameters (M) & Memory (GB) & Iterations per second \\ \hline \hline
FNO          &  2.376             & 12.74           &  3.45              \\ \hline
DeepONet     &  0.297             &  -              &  -                 \\ \hline
ResNet       &  \textbf{0.148}             & 4.66            &  9.25               \\ \hline
ViTO         &  0.150             & \textbf{0.85}            &  \textbf{59.98}             \\ \hline
\end{tabular}
\caption{Computational performance of the different models.}
\label{details}
\end{table}

In Table \ref{details} we present a computational comparison of the four models mentioned above. These results are given for a Darcy problem (see \ref{darcy_problem}) experiment with a grid size of $128 \times 128$ and batch size 100. Memory was calculated as the peak GPU memory usage from the beginning of the training process to its end. The iterations per second metric was calculated by measuring the time it took the model to train for a single batch, averaged over 200 batches to increase consistency. Note that we do not report these two metrics for the DeepONet, since the training process was quite different from the other three models, so any direct comparison would have been misleading.

ViTO was the most efficient model by a substantial margin, both in terms of memory consumption and training time. ViTO was able to produce results on-par with SOTA methods, using a surprisingly small number of trainable parameters. The full details of the ViT-related weights in ViTO are shown in Table \ref{Vit_architecture}. Despite having a similar number of parameters to the ResNet, ViTO employs downsampling due to its U-Net architecture. Consequently, many convolutional operations occur on smaller feature maps compared to the ResNet which explains the lower memory usage and running time for ViTO. 

\begin{table}[!ht]
\centering
\begin{tabular}{|c|c|c|c|c|}
\hline
Problem       & Transformer blocks & Attention heads & Embedding dimension & ViT MLP size \\ \hline \hline
Wave equation &   2                &    2            &      16             &      128      \\ \hline
Navier-Stokes &   4                &    8            &      16             &      64      \\ \hline
Darcy Flow    &   2                &    2            &      16             &      128     \\ \hline
\end{tabular}
\caption{ViT parameters for each scenario.}
\label{Vit_architecture}
\end{table}

\subsection{Wave equation}\label{wave_problem}

The formulation of the acoustic wave equation in two dimensions is given by \cite{evans2022partial, jost2012partial}:

\begin{equation}
	\begin{cases}
		\ddot{u}(x, y, t)=c^2(x,y) (u_{xx}(x, y, t) + u_{yy}(x, y, t)) + f(x, y, t) & (x,y) \in (0,L)^2; 0\leq t \leq T, \\
		u(x, y, 0)=u_0(x, y) & (x,y) \in (0,L)^2, \\
		\dot{u}(x, y, 0)=v_0(x,y) & (x,y) \in (0,L)^2, \\
		u(0, y, t)=u(L, y, t)=0 & y \in (0,L),\; 0\leq t \leq T, \\
		u(x, 0, t)=u(x, L, t)=0 & x \in (0,L),\; 0\leq t \leq T. \\
	\end{cases}
\end{equation}

where $u(x,y,t)$ is the wave amplitude or acoustic pressure, $c(x,y)$ is the wave propagation speed, $f(x,y,t)$ is the source term, $T$ is the final propagation time, $L$ is the size of the physical domain, and $u_0(x, y), v_0(x, y)$ are the initial pressures and velocities, respectively. The boundary condition is a homogeneous Dirichlet boundary (fully-reflective). The inverse problem is to learn the following mapping:

\begin{equation*}
    u(x,y,T) \longmapsto u_0(x,y).
\end{equation*}
 
We chose a physical domain with $L = \pi$ and propagation time $T = 0.001$. We set the initial pressure and velocity to be 0, and randomly created Gaussian-shaped sources at different locations. For each sample, we created two such Gaussian sources with random amplitudes and locations. The locations are selected using a discrete random uniform distribution on the indices of the grid. The amplitudes are sampled uniformly using $\mathcal{U}(-1, 1)$ for each source in each initial condition. The wave velocity was taken as $c(x,y) = c_0 sin(x) sin(y)$, where $c_0$ was randomly sampled for each initial condition using a uniform distribution set between 1,300 and 1,600, which is centred around the average acoustic wave propagation speed of 1,484 in the Mediterranean sea.  

The dataset was generated using a standard explicit second-order finite-difference scheme \cite{hyper1, hyper2}. We generated 20,000 samples, of which 16,000 were used as the training set, while the remaining 4,000 were evenly split to form the testing and validation sets.

\begin{figure}[htb]
\includegraphics[scale=0.35]{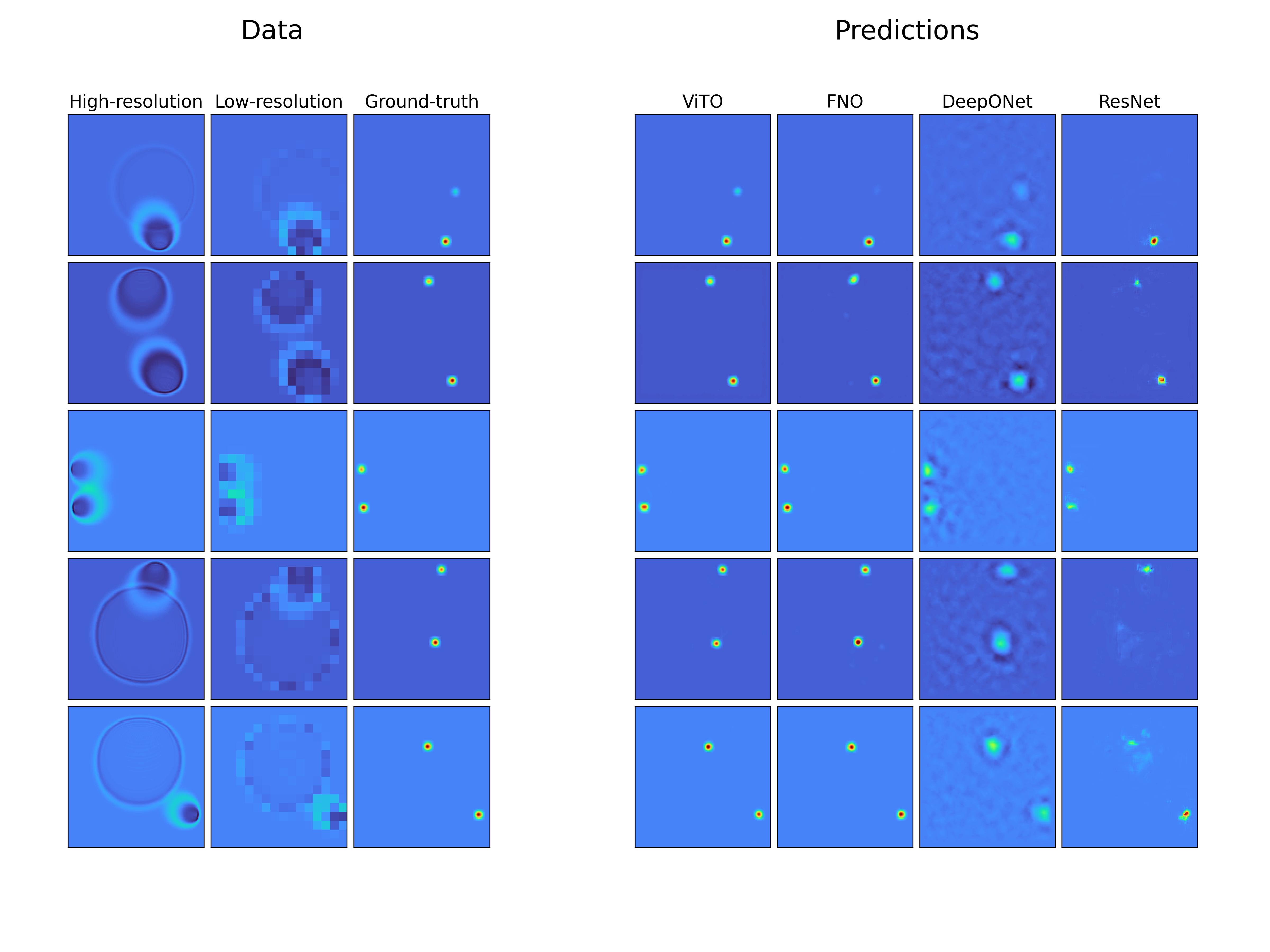}
\centering
\caption{Predictions for the wave problem \ref{wave_problem} for 5 random samples. In the section labeled "Data" is the propagated wave at the final time using fine and coarse discretizations, alongside a high-resolution image of the ground truth sources. In the section labeled "Predictions" are the initial condition reconstructions by the various methods.}
\label{fig:wave}
\end{figure}

The results are shown in Table \ref{tab:wave_results} and Figure \ref{fig:wave}. ViTO obtains the lowest error compared to the other methods, both with and without noise. It is worth noting that ViTO is able to reconstruct the initial condition even in difficult scenarios where there is a large difference between the amplitudes of the different sources (such as the first row in Figure \ref{fig:wave}).  

\begin{table}[!ht]
\centering
\begin{tabular}{|c|c|c|c|c|}
\hline
Method       & $0 \%$ noise       & $10 \%$ noise   \\ \hline \hline
FNO          & 0.3260             & 0.4383          \\ \hline
DeepONet     & 0.7128             & 0.7131          \\ \hline
ResNet       & 0.7892             & 0.8154          \\ \hline
ViTO         & \textbf{0.2678}             & \textbf{0.2942}         \\ \hline
\end{tabular}
\caption{Test relative $\textit{L}^2$ errors for the wave problem.}
\label{tab:wave_results}
\end{table}

\subsection{Navier-Stokes equations}\label{ns_problem}

The time-dependent two-dimensional Navier-Stokes equation for the viscous, incompressible fluid in vorticity form is given by:

\begin{equation}\label{NS_equation}
\begin{cases}
    \partial_t \omega(x,y,t) + u(x,y,t) \cdot \nabla  \omega(x,y,t) = \nu \Delta \omega(x,y,t) + f(x,y), & x,y \in (0,1)^2, t \in (0, T] \\
    \nabla \cdot u(x,y,t) = 0, & (x,y) \in (0,1)^2, t \in (0, T] \\
    \omega(x,y,0) = \omega_0, & (x,y) \in (0,1)^2 \\
\end{cases}
\end{equation}
where $\omega$ is the vorticity, $u$ is the velocity field,  $\nu = 10^{-3}$ is the viscosity, and $\Delta$ is the two-dimensional Laplacian. We consider periodic boundary conditions. The source term $f$ is set as: $f(x,y) = 0.1 (sin(2\pi (x + y)) + cos(2\pi(x + y)))$, and the initial condition $\omega_0(x)$ is sampled from a Gaussian random field according to the following distribution: $\mathcal{N}(0, 7^{3/2} (-\Delta + 49 I)^{-5/2})$. The inverse problem is to learn the following mapping:

\begin{equation*}
    \omega(x,y,T) \longmapsto \omega_0(x,y).
\end{equation*}

 We used the publicly available Python solver given in \cite{FNO} to create two separate datasets with different final simulation times $T=1$ and $T=5$. Each dataset was composed of 10,000 samples, which we then split into train, test, and validation. The error analysis for $T=1,5$ are shown in Table \ref{tab:ns}. ViTO and FNO obtain very similar accuracy in both cases,  ViTO is slightly more accurate without noise, while FNO has a minor advantage with noise. 
 
\begin{figure}[htb]
\includegraphics[scale=0.35]{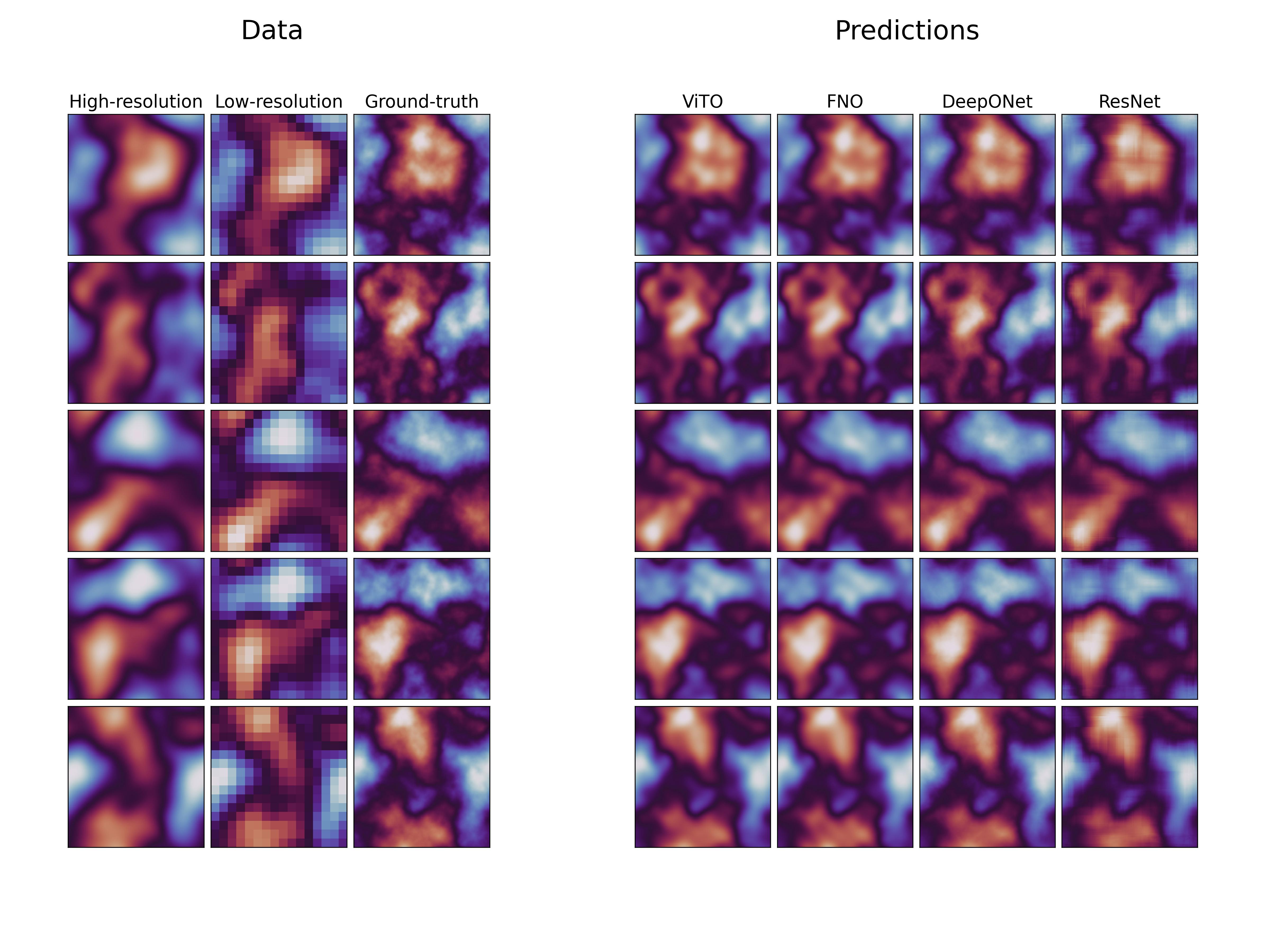}
\centering
\caption{Predictions for the Navier-Stokes problem \ref{ns_problem} with $T=1$ for 5 random samples. In the section labeled "Data" is the vorticity at the final time using fine and coarse discretizations, alongside a high-resolution image of the initial vorticity. In the section labeled "Predictions" are the initial condition reconstructions by the various methods.}
\label{fig:ns_results_1}
\end{figure}

\begin{figure}[htb]
\includegraphics[scale=0.35]{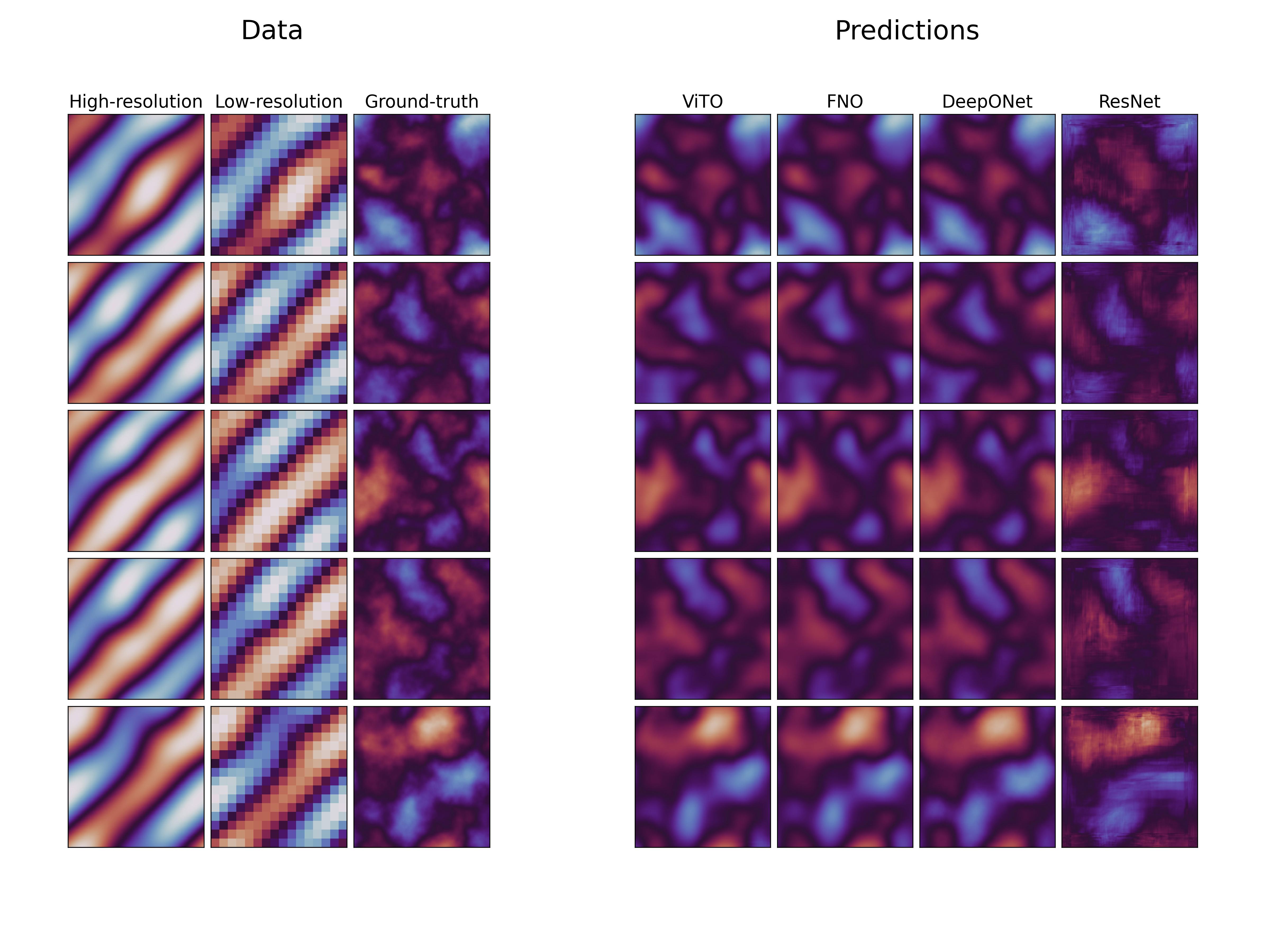}
\centering
\caption{Predictions for the Navier-Stokes problem \ref{ns_problem} with $T=5$ for 5 random samples. In the section labeled "Data" is the vorticity at the final time using fine and coarse discretizations, alongside a high-resolution image of the initial vorticity. In the section labeled "Predictions" are the initial condition reconstructions by the various methods.}
\label{fig:ns_results_5}
\end{figure}

 A visualization of the results is shown in Figure \ref{fig:ns_results_1} and Figure \ref{fig:ns_results_5}. In the case $T=1$ (Figure \ref{fig:ns_results_1}), the vorticities for different initial conditions are still very different from one another, and so the reconstructions are able to capture fine details. However, for $T=5$ (Figure \ref{fig:ns_results_5}), the behavior of the vorticity becomes very similar, regardless of the choice of initial condition. In that case, some fine details are lost in all reconstructions, which explains the larger error compared to the $T=1$ case.

\begin{table}[ht]\caption{Test relative $\textit{L}^2$ errors for the errors for the Navier-Stokes problem with different final simulation times ($T$) and noise levels ($\gamma$).}
\centering 
\begin{tabular}{ccccccc} %
\toprule
\multirow{2}{*}{
\parbox[c]{.2\linewidth}{}}
  & \multicolumn{2}{c}{$T=1$} &&
\multicolumn{2}{c}{$T=5$} \\ 
\cmidrule{2-3} \cmidrule{5-6}

 & {\centering $\gamma = 0$} & {$\gamma = 0.1$} && {$\gamma = 0$} & {$\gamma = 0.1$} \\
\midrule
FNO      & 0.06449 & \textbf{0.1587} &&   0.1881  & \textbf{0.3582} \\
DeepONet & 0.09424 & 0.1684 &&   0.2007  & 0.4528 \\
ResNet   & 0.1271 & 0.4471 &&   0.4520  & 0.5745 \\
ViTO     & \textbf{0.06348} & 0.1635 &&   \textbf{0.1757} & 0.3757 \\
\bottomrule
\end{tabular}
\label{tab:ns}
\end{table}


\subsection{Darcy equation}\label{darcy_problem}
The steady-state two-dimensional Darcy flow for a porous medium is given by the following equation:
\begin{equation}\label{darcy_equation}
\begin{cases}
    -\nabla \cdot (K(x,y) \nabla h(x,y))= f(x,y), & (x,y) \in (0,1)^2  \\
    h(x,y) = 0, & (x,y) \in \partial (0,1)^2 \\
\end{cases}
\end{equation}

where $\partial (0,1)^2$ is the domain boundary, $K(x, y)$ is the permeability coefficient field, $h(x, y)$ is the pressure, and $f$ is a forcing function. The boundary condition used here is a homogeneous Dirichlet boundary (fully-reflective). The inverse problem is to learn the following mapping:

\begin{equation*}
    h(x,y) \longmapsto K(x,y).
\end{equation*}

We used the publicly available finite difference solver (written in MATLAB \cite{MATLAB}, given in \cite{FNO}) to create data with piecewise smooth coefficients $K$, with a constant forcing function $f \equiv 1$. The coefficient was selected using a Gaussian random field according to the following distribution: $\mathcal{N}(0, (-\Delta + 9 I)^{-2})$. This is followed by a binarization operation that mapped positive values to 12 and negative values to 3. We created 3 such datasets for the following resolutions: $n=128, 256, 512$ with a SR scale factor of 8. Hence, the super-resolution mappings were of the following dimensions:  $16 \times 16 \longmapsto 128 \times 128$, $32 \times 32 \longmapsto 256 \times 256$, and $64 \times 64 \longmapsto 512 \times 512$. Each dataset contained $1,000$ samples which were split into $800, 100, 100$ training, validation, and testing samples, respectively. Despite being a binary problem, we still used the $\textit{L}^2$ loss function \eqref{eqn:loss}, and not a binary loss function like negative log likelihood, since $K(x,y)$ does not have to be binary in many applications. For the two datasets with finer grids we had to use a smaller batch size of 10 to fit the models into memory. We note that ViTO was the only model we were able to run with a batch size of 100, but we kept it at 10 to make the comparison more accurate. 

The full error analysis is shown in Table \ref{tab:darcy}. In all 6 cases ViTO obtained the best accuracy compared to the benchmark methods. Note that the results for the Darcy problem and the wave problem \ref{wave_problem} are more decisive in comparison to the Navier-Stokes problem \ref{ns_problem}. This could potentially be explained by the shift from smooth functions to functions consisting of irregular interfaces and sharp features. Recall that FNOs rely on Fourier transforms, which can be very accurate for smooth functions, but face severe difficulties with discontinuities. Furthermore, as we refined the grid in the Darcy case we saw a noticeable improvement in the ViTO results, which was not observed in the FNO case. This can also be explained by Fourier analysis, since FNOs are learning a global base of functions, which renders them grid-invariant. 

\begin{figure}
\centering
\begin{subfigure}[b]{0.9\textwidth}
   \includegraphics[scale=0.36]{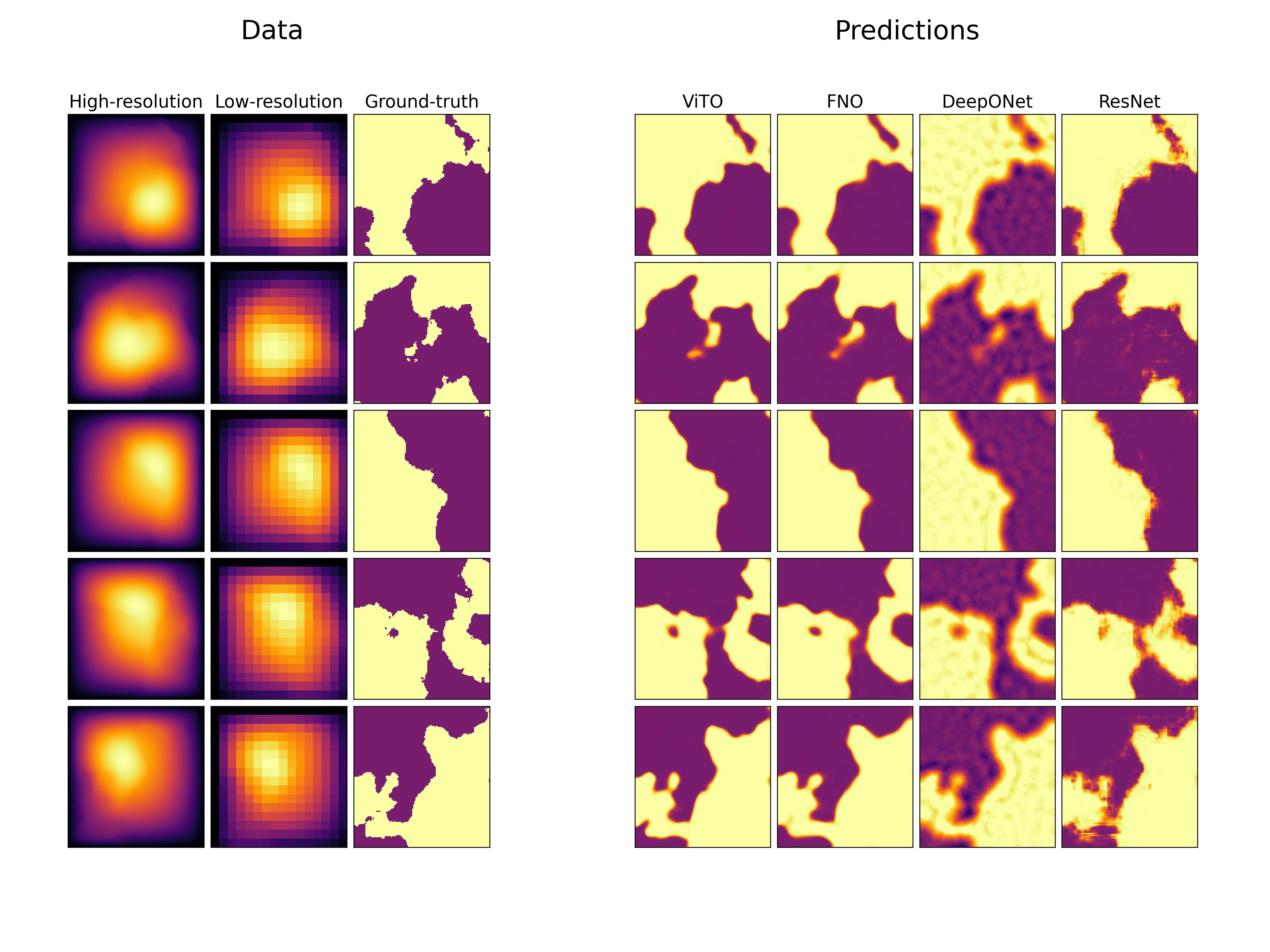}
   \caption{Results for $n=128$.}
   \label{fig:darcy128} 
\end{subfigure}
\begin{subfigure}[b]{0.9\textwidth}
   \includegraphics[scale=0.36]{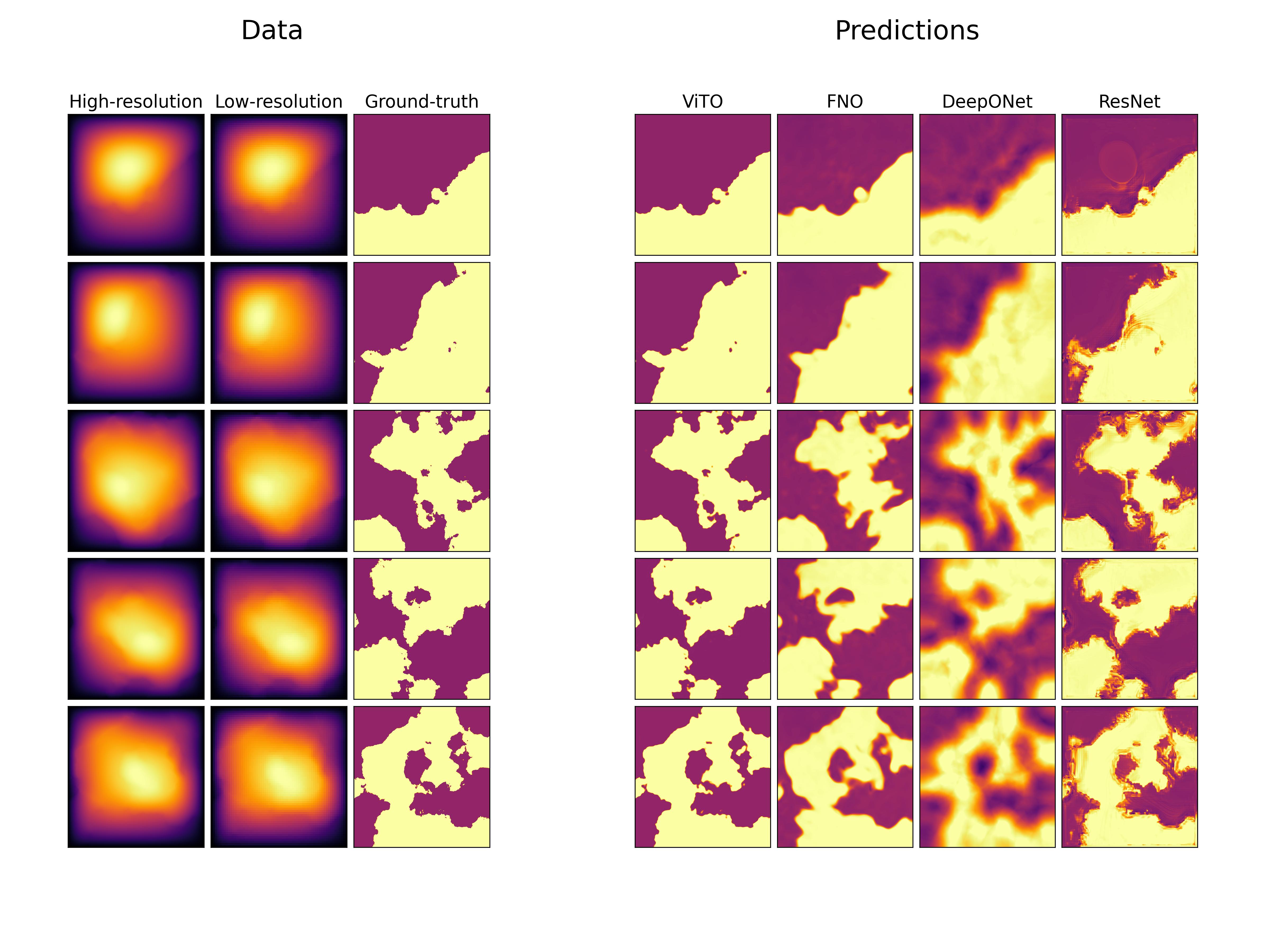}
   \caption{Results for $n=512$.}
   \label{fig:darcy512}
\end{subfigure}
\caption{Predictions for the Darcy problem \ref{darcy_problem} for 5 random samples using for $n=128$ and $n=512$. In the section labeled "Data" is the PDE solution $u(x,y)$ using fine and coarse discretizations, alongside a high-resolution image of the permeability  coefficient field  $K(x,y)$. In the section labeled "Predictions" are the permeability coefficient reconstructions by the various methods.}
\label{fig:darcy}
\end{figure}

Visualizations of the results for $n=128$ and $n=512$ are shown in Figure \ref{fig:darcy}, respectively. Note that ViTO was able to capture sharp features of the coefficient $K(x,y)$. This is especially noticeable in cases where there are very small discontinuities (cavities) in the data; while ViTO was mostly able to capture them, FNO tended to smooth them.


\begin{table}[ht]\caption{Test relative $\textit{L}^2$ errors for the Darcy flow problem with different grid sizes ($n$) and noise levels ($\gamma$).}
\centering 
\begin{tabular}{ccccccccc} %
\toprule
\multirow{2}{*}{
\parbox[c]{.2\linewidth}{}}
  & \multicolumn{2}{c}{$n=128$} &&
\multicolumn{2}{c}{$n=256$} &&
\multicolumn{2}{c}{$n=512$} \\ 
\cmidrule{2-3} \cmidrule{5-6} \cmidrule{8-9} 

 & {\centering $\gamma = 0$} & {$\gamma = 0.1$} && {$\gamma = 0$} & {$\gamma = 0.1$}  && {$\gamma = 0$} & {$\gamma = 0.1$} \\
\midrule
FNO      & 0.1422 & 0.4502 &&   0.1272  & 0.1915  &&  0.1235  & 0.1683 \\
DeepONet & 0.1463 & 0.4502 &&   0.1422  & 0.2090  &&  0.1608  & 0.2174 \\
ResNet   & 0.1603 & 0.2760 &&   0.1287  & 0.3078  &&  0.1416  & 0.3702 \\
ViTO     & \textbf{0.1184} & \textbf{0.1943} &&   \textbf{0.08216} & \textbf{0.1799}  &&  \textbf{0.05197} & \textbf{0.1623} \\
\bottomrule
\end{tabular}
\label{tab:darcy}
\end{table}

\subsection{Varying input size}\label{var_input}

Finally, we assessed the ability of ViTO to handle inputs of various sizes without requiring retraining. Typically, transformers are capable of handling such inputs, which are prevalent in NLP contexts. As mentioned in \ref{network_arch}, we employed relative positional encoding, which was shown to be effective for computer vision problems of this sort. Additionally, the U-Net architecture is fully convolutional \cite{long2015fully}, allowing it to handle such inputs.

To evaluate this capability, we used the Darcy example \ref{darcy_problem} with $n=512$. We followed the same steps as in all other experiments, with one addition to the training process. During each training batch, we randomly selected a subsampling parameter $r \in \{1, 2, 3, \hdots , 9 \}$ and applied it to the input image (rounding the number of grid points to the nearest integer). For instance, taking $r =4$, an input of size $512 \times 512$ was downsampled to size $128 \times 128$. We used this process to allow ViTO to generalize better for new discretizations. This procedure can be considered a type of data augmentation. We also dropped the super-resolution part of the inverse problem (i.e. $s=1$ in \ref{data_driven}) to enable us to run tests for large grids.

\begin{figure}[!htb]
\centering
\begin{subfigure}[b]{0.9\textwidth}
   \includegraphics[scale=0.35]{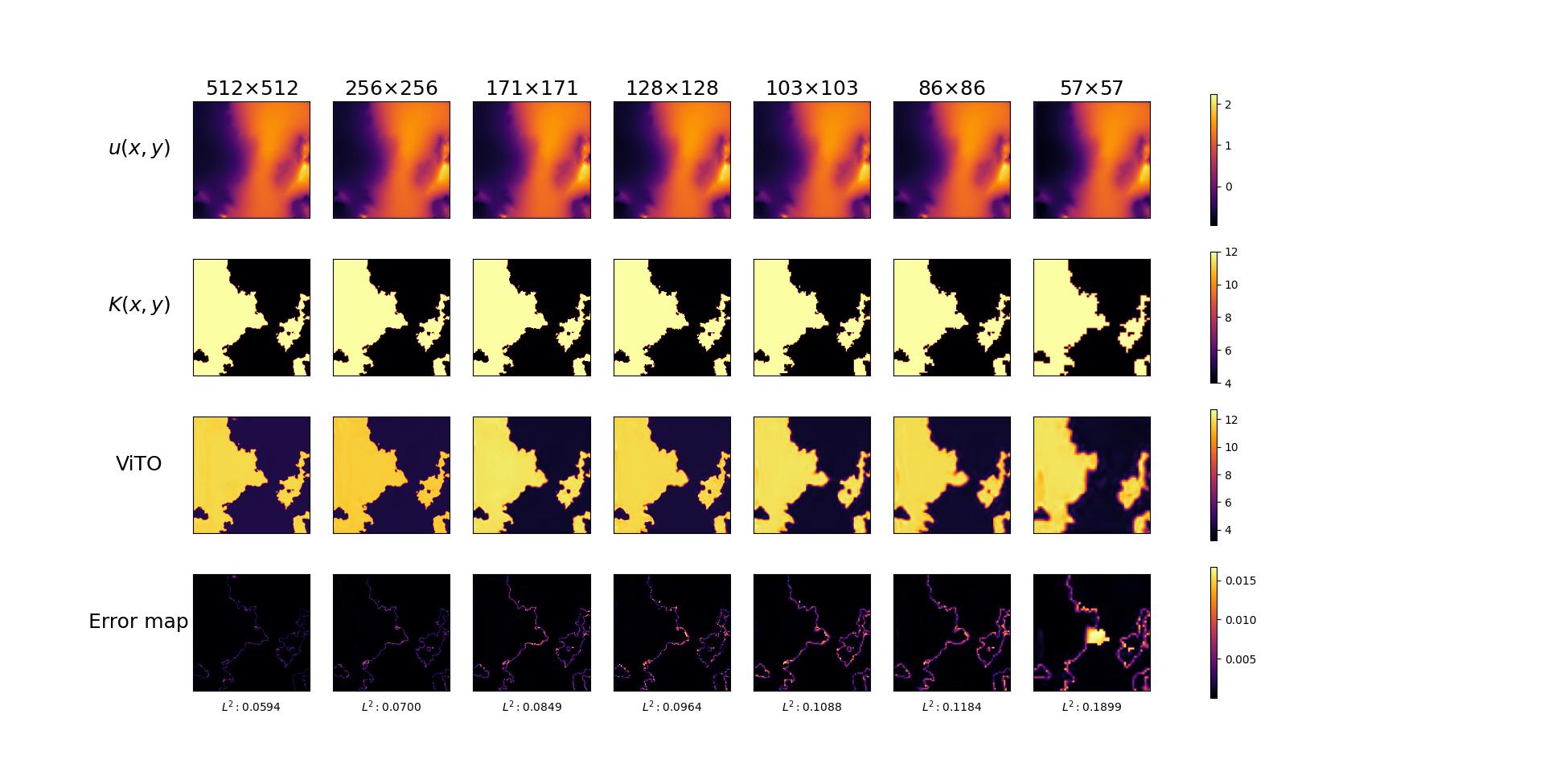}
   \caption{Previously seen discretizations.}
   \label{fig:variable_grid} 
\end{subfigure}
\begin{subfigure}[b]{0.9\textwidth}
   \includegraphics[scale=0.35]{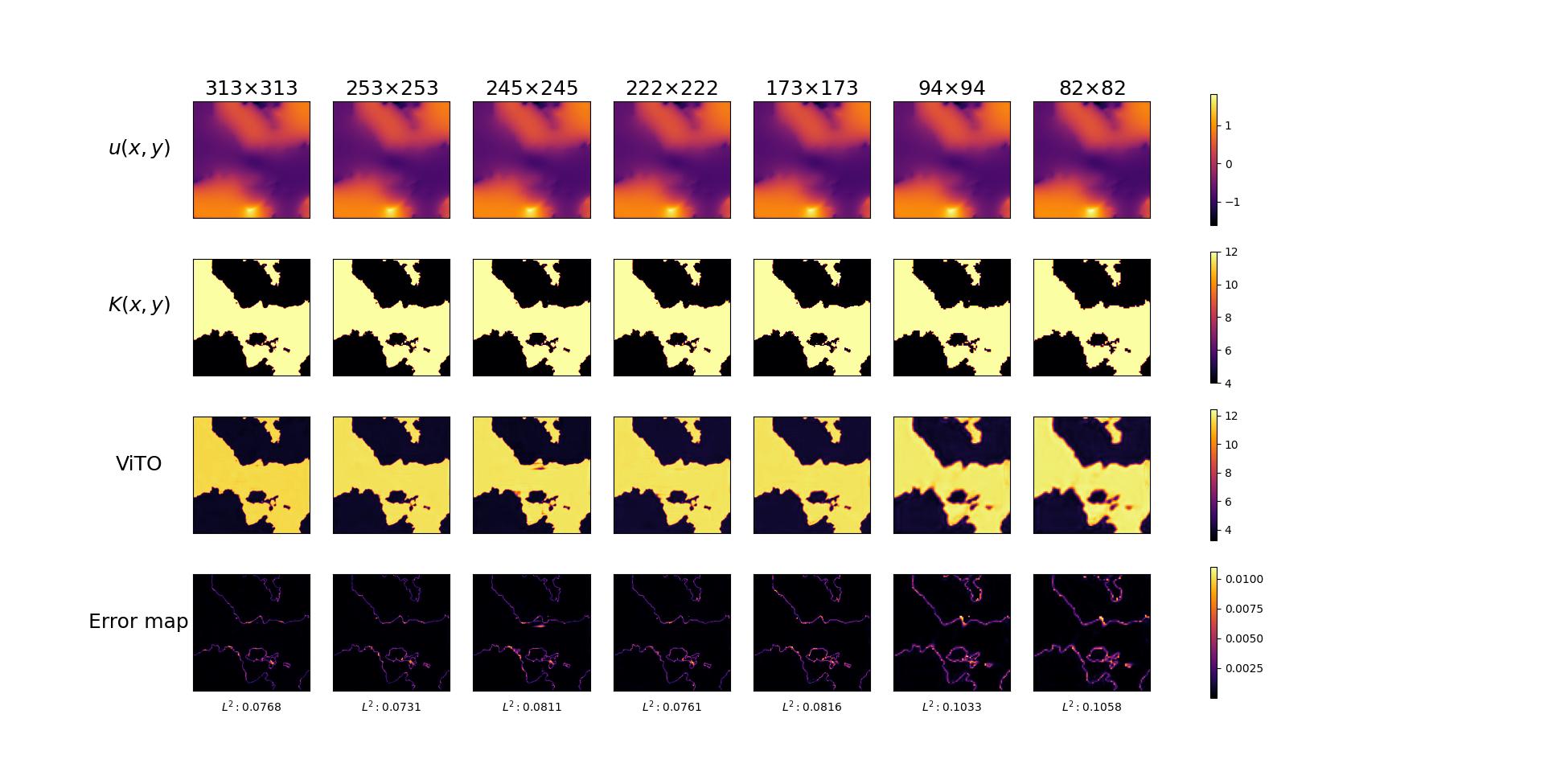}
   \caption{Zero-shot discretizations.}
   \label{fig:zero_shot_variable_grid}
\end{subfigure}
\caption{ViTO predictions with varying input sizes. The results are presented in columns corresponding to different discretizations. The first two rows show the ground-truth values of $u(x,y)$ and $K(x,y)$ as given by \eqref{darcy_equation}. The third row presents the predictions of ViTO. The last row shows the point-wise relative $\textit{L}^2$ error between $K(x,y)$ and the ViTO prediction.}
\label{fig:variable_input}
\end{figure}

Finally, we tested the model twice. First, we evaluated it on samples from the test set with grid sizes it had encountered during training, which were: $\{ \frac{512}{r}: r = 1, \hdots , 9 \}$, rounded to the nearest integer. Next, we created a zero-shot scenario, where the model was presented with samples having random discretizations that it had not seen during training. We created these discretizations by resizing the original samples accordingly. The results are presented in Figure \ref{fig:variable_input}. The results show that ViTO is capable of handling different grids without retraining, even in a zero-shot scenario. The error maps show us that larger grids generally yield better results, and that the error is mostly concentrated around the discontinuities (due to the binarization of the permeability field).

\newpage

\section{Discussion and future work}\label{discussion}

We have introduced a novel approach to inverse problems and super-resolution, which incorporates vision transformers with operator learning. Our approach, named ViTO, combines
a U-Net based architecture with a vision transformer. We have obtained comparable or superior results compared to the leading operator network benchmarks
in terms of accuracy, accompanied by substantial efficiency gains.

The impressive performance of ViTO on inverse problems of considerable complexity requires thorough investigation to uncover the mathematical and algorithmic reasons behind it. In particular, we should understand the learning mechanism in the latent space and provide some theoretical background.

Extending ViTO to solve  problems with three spatial dimensions is an avenue worth exploring. Moreover, we need to determine ViTO's ability to adapt to forward problems, especially those that are time-dependent.

\section{Acknowledgements}

This work was supported by the Vannevar Bush Faculty Fellowship award (GEK) from ONR (N00014-22-1-2795).
The work of PS and GEK is supported by the U.S. Department of Energy, Advanced Scientific Computing Research program, under the Scalable, Efficient and Accelerated Causal Reasoning Operators, Graphs and Spikes for Earth and Embedded Systems (SEA-CROGS) project, DE-SC0023191. Pacific Northwest National Laboratory (PNNL) is a multi-program national laboratory operated for the U.S. Department of Energy (DOE) by Battelle Memorial Institute under Contract No. DE-AC05-76RL01830.

\bibliographystyle{unsrt}  
\bibliography{references}

\begin{thebibliography}{10}

\bibitem{deeponet}
Lu~Lu, Pengzhan Jin, Guofei Pang, Zhongqiang Zhang, and George~Em Karniadakis.
\newblock Learning nonlinear operators via {DeepONet} based on the universal
  approximation theorem of operators.
\newblock {\em Nature Machine Intelligence}, 3(3):218--229, 2021.

\bibitem{FNO}
Zongyi Li, Nikola Kovachki, Kamyar Azizzadenesheli, Burigede Liu, Kaushik
  Bhattacharya, Andrew Stuart, and Anima Anandkumar.
\newblock Fourier neural operator for parametric partial differential
  equations.
\newblock {\em arXiv preprint arXiv:2010.08895}, 2020.

\bibitem{FNO2}
Nikola~B. Kovachki, Zongyi Li, Burigede Liu, Kamyar Azizzadenesheli, Kaushik
  Bhattacharya, Andrew~M. Stuart, and Anima Anandkumar.
\newblock Neural operator: Learning maps between function spaces.
\newblock {\em CoRR}, abs/2108.08481, 2021.

\bibitem{transformer}
Ashish Vaswani, Noam Shazeer, Niki Parmar, Jakob Uszkoreit, Llion Jones,
  Aidan~N. Gomez, Lukasz Kaiser, and Illia Polosukhin.
\newblock Attention is all you need.
\newblock {\em CoRR}, abs/1706.03762, 2017.

\bibitem{vit}
Alexey Dosovitskiy, Lucas Beyer, Alexander Kolesnikov, Dirk Weissenborn,
  Xiaohua Zhai, Thomas Unterthiner, Mostafa Dehghani, Matthias Minderer, Georg
  Heigold, Sylvain Gelly, Jakob Uszkoreit, and Neil Houlsby.
\newblock An image is worth 16x16 words: Transformers for image recognition at
  scale.
\newblock {\em CoRR}, abs/2010.11929, 2020.

\bibitem{imagenet}
Jia Deng, Wei Dong, Richard Socher, Li-Jia Li, Kai Li, and Li~Fei-Fei.
\newblock Imagenet: A large-scale hierarchical image database.
\newblock In {\em 2009 IEEE conference on computer vision and pattern
  recognition}, pages 248--255. IEEE, 2009.

\bibitem{cifar}
Alex Krizhevsky, Geoffrey Hinton, et~al.
\newblock Learning multiple layers of features from tiny images, 2009.

\bibitem{oxford_pets}
Omkar~M. Parkhi, Andrea Vedaldi, Andrew Zisserman, and C.~V. Jawahar.
\newblock Cats and dogs.
\newblock In {\em IEEE Conference on Computer Vision and Pattern Recognition},
  2012.

\bibitem{VTAB}
Xiaohua Zhai, Joan Puigcerver, Alexander Kolesnikov, Pierre Ruyssen, Carlos
  Riquelme, Mario Lucic, Josip Djolonga, Andr{\'{e}}~Susano Pinto, Maxim
  Neumann, Alexey Dosovitskiy, Lucas Beyer, Olivier Bachem, Michael Tschannen,
  Marcin Michalski, Olivier Bousquet, Sylvain Gelly, and Neil Houlsby.
\newblock The visual task adaptation benchmark.
\newblock {\em CoRR}, abs/1910.04867, 2019.

\bibitem{okolo2022ievit}
Gabriel~Iluebe Okolo, Stamos Katsigiannis, and Naeem Ramzan.
\newblock Ievit: An enhanced vision transformer architecture for chest x-ray
  image classification.
\newblock {\em Computer Methods and Programs in Biomedicine}, 226:107141, 2022.

\bibitem{chen2021vit}
Junyu Chen, Yufan He, Eric~C Frey, Ye~Li, and Yong Du.
\newblock Vit-v-net: Vision transformer for unsupervised volumetric medical
  image registration.
\newblock {\em arXiv preprint arXiv:2104.06468}, 2021.

\bibitem{transformerPDE1}
Zijie Li, Kazem Meidani, and Amir~Barati Farimani.
\newblock Transformer for partial differential equations' operator learning.
\newblock {\em arXiv preprint arXiv:2205.13671}, 2022.

\bibitem{transformerPDE2}
Xinliang Liu, Bo~Xu, and Lei Zhang.
\newblock Ht-net: Hierarchical transformer based operator learning model for
  multiscale pdes.
\newblock {\em arXiv preprint arXiv:2210.10890}, 2022.

\bibitem{transformerPDE3}
Zhongkai Hao, Chengyang Ying, Zhengyi Wang, Hang Su, Yinpeng Dong, Songming
  Liu, Ze~Cheng, Jun Zhu, and Jian Song.
\newblock Gnot: A general neural operator transformer for operator learning.
\newblock {\em arXiv preprint arXiv:2302.14376}, 2023.

\bibitem{transformer_cao_choose}
Shuhao Cao.
\newblock Choose a transformer: Fourier or galerkin.
\newblock {\em Advances in neural information processing systems},
  34:24924--24940, 2021.

\bibitem{transformer_cao2}
Ruchi Guo, Shuhao Cao, and Long Chen.
\newblock Transformer meets boundary value inverse problems.
\newblock {\em arXiv preprint arXiv:2209.14977}, 2022.

\bibitem{ronneberger2015u}
Olaf Ronneberger, Philipp Fischer, and Thomas Brox.
\newblock U-net: Convolutional networks for biomedical image segmentation.
\newblock In {\em Medical Image Computing and Computer-Assisted
  Intervention--MICCAI 2015: 18th International Conference, Munich, Germany,
  October 5-9, 2015, Proceedings, Part III 18}, pages 234--241. Springer, 2015.

\bibitem{chen2021transunet}
Jieneng Chen, Yongyi Lu, Qihang Yu, Xiangde Luo, Ehsan Adeli, Yan Wang, Le~Lu,
  Alan~L Yuille, and Yuyin Zhou.
\newblock Transunet: Transformers make strong encoders for medical image
  segmentation.
\newblock {\em arXiv preprint arXiv:2102.04306}, 2021.

\bibitem{deeponet-fno}
Lu~Lu, Xuhui Meng, Shengze Cai, Zhiping Mao, Somdatta Goswami, Zhongqiang
  Zhang, and George~Em Karniadakis.
\newblock A comprehensive and fair comparison of two neural operators (with
  practical extensions) based on fair data.
\newblock {\em Computer Methods in Applied Mechanics and Engineering},
  393:114778, 2022.

\bibitem{hadamard1902problemes}
Jacques Hadamard.
\newblock Sur les probl{\`e}mes aux d{\'e}riv{\'e}es partielles et leur
  signification physique.
\newblock {\em Princeton university bulletin}, pages 49--52, 1902.

\bibitem{ResNet}
Kaiming He, Xiangyu Zhang, Shaoqing Ren, and Jian Sun.
\newblock Deep residual learning for image recognition.
\newblock In {\em Proceedings of the IEEE conference on computer vision and
  pattern recognition}, pages 770--778, 2016.

\bibitem{batch_norm}
Sergey Ioffe and Christian Szegedy.
\newblock Batch normalization: Accelerating deep network training by reducing
  internal covariate shift.
\newblock In {\em International conference on machine learning}, pages
  448--456. pmlr, 2015.

\bibitem{hendrycks2016gaussian}
Dan Hendrycks and Kevin Gimpel.
\newblock Gaussian error linear units (gelus).
\newblock {\em arXiv preprint arXiv:1606.08415}, 2016.

\bibitem{shaw2018self}
Peter Shaw, Jakob Uszkoreit, and Ashish Vaswani.
\newblock Self-attention with relative position representations.
\newblock {\em arXiv preprint arXiv:1803.02155}, 2018.

\bibitem{wu2021rethinking}
Kan Wu, Houwen Peng, Minghao Chen, Jianlong Fu, and Hongyang Chao.
\newblock Rethinking and improving relative position encoding for vision
  transformer.
\newblock In {\em Proceedings of the IEEE/CVF International Conference on
  Computer Vision}, pages 10033--10041, 2021.

\bibitem{chollet2017xception}
Fran{\c{c}}ois Chollet.
\newblock Xception: Deep learning with depthwise separable convolutions.
\newblock In {\em Proceedings of the IEEE conference on computer vision and
  pattern recognition}, pages 1251--1258, 2017.

\bibitem{chu2021conditional}
Xiangxiang Chu, Zhi Tian, Bo~Zhang, Xinlong Wang, Xiaolin Wei, Huaxia Xia, and
  Chunhua Shen.
\newblock Conditional positional encodings for vision transformers.
\newblock {\em arXiv preprint arXiv:2102.10882}, 2021.

\bibitem{ill_posed}
Andrei~Nikolaevich Tikhonov, AV~Goncharsky, Vyacheslav~Vasil'evich Stepanov,
  and Anatoly~G Yagola.
\newblock {\em Numerical methods for the solution of ill-posed problems},
  volume 328.
\newblock Springer Science \& Business Media, 1995.

\bibitem{urban100}
Jia-Bin Huang, Abhishek Singh, and Narendra Ahuja.
\newblock Single image super-resolution from transformed self-exemplars.
\newblock In {\em Proceedings of the IEEE Conference on Computer Vision and
  Pattern Recognition (CVPR)}, June 2015.

\bibitem{BSD}
D.~Martin, C.~Fowlkes, D.~Tal, and J.~Malik.
\newblock A database of human segmented natural images and its application to
  evaluating segmentation algorithms and measuring ecological statistics.
\newblock In {\em Proc. 8th Int'l Conf. Computer Vision}, volume~2, pages
  416--423, July 2001.

\bibitem{kingma2014adam}
Diederik~P Kingma and Jimmy Ba.
\newblock Adam: A method for stochastic optimization.
\newblock {\em arXiv preprint arXiv:1412.6980}, 2014.

\bibitem{adamw}
Ilya Loshchilov and Frank Hutter.
\newblock Decoupled weight decay regularization.
\newblock {\em arXiv preprint arXiv:1711.05101}, 2017.

\bibitem{loshchilov2016sgdr}
Ilya Loshchilov and Frank Hutter.
\newblock Sgdr: Stochastic gradient descent with warm restarts.
\newblock {\em arXiv preprint arXiv:1608.03983}, 2016.

\bibitem{pytorch}
Adam Paszke, Sam Gross, Francisco Massa, Adam Lerer, James Bradbury, Gregory
  Chanan, Trevor Killeen, Zeming Lin, Natalia Gimelshein, Luca Antiga, et~al.
\newblock Pytorch: An imperative style, high-performance deep learning library.
\newblock {\em Advances in neural information processing systems}, 32, 2019.

\bibitem{deepxde}
Lu~Lu, Xuhui Meng, Zhiping Mao, and George~Em Karniadakis.
\newblock {DeepXDE}: A deep learning library for solving differential
  equations.
\newblock {\em SIAM Review}, 63(1):208--228, 2021.

\bibitem{evans2022partial}
Lawrence~C Evans.
\newblock {\em Partial differential equations}, volume~19.
\newblock American Mathematical Society, 2022.

\bibitem{jost2012partial}
J{\"u}rgen Jost.
\newblock {\em Partial differential equations}, volume 214.
\newblock Springer Science \& Business Media, 2012.

\bibitem{hyper1}
R.~Abgrall and C.W. Shu.
\newblock {\em Handbook of Numerical Methods for Hyperbolic Problems: Basic and
  Fundamental Issues}.
\newblock Elsevier, North Holland, 1 edition, 2016.

\bibitem{hyper2}
J.~Oliger B.~Gustafsson, H.O.~Kreiss.
\newblock {\em Time-Dependent Problems and Difference Methods}.
\newblock Pure and Applied Mathematics, Wiley, 2 edition, 2016.

\bibitem{MATLAB}
The~MathWorks Inc.
\newblock Matlab version: 9.13.0 (r2022b), 2022.

\bibitem{long2015fully}
Jonathan Long, Evan Shelhamer, and Trevor Darrell.
\newblock Fully convolutional networks for semantic segmentation.
\newblock In {\em Proceedings of the IEEE conference on computer vision and
  pattern recognition}, pages 3431--3440, 2015.

\end{thebibliography}

\end{document}